\documentclass[letterpaper]{article} 
\usepackage{aaai2026}  
\usepackage{times}  
\usepackage{helvet}  
\usepackage{courier}  
\usepackage[hyphens]{url}  
\usepackage{graphicx} 
\usepackage{natbib}  
\usepackage{caption} 
\usepackage{algorithm}
\usepackage{algorithmic}
\usepackage{pdfpages}

\usepackage{newfloat}
\usepackage{listings}
\DeclareCaptionStyle{ruled}{labelfont=normalfont,labelsep=colon,strut=off} 
\floatstyle{ruled}
\newfloat{listing}{tb}{lst}{}
\floatname{listing}{Listing}
\usepackage{amsmath}
\usepackage{bm}
\usepackage{amssymb}   
\usepackage{multirow}  
\usepackage{booktabs}  
\nocopyright

\title{Learning by Imagining: Debiased Feature Augmentation for \\ Compositional Zero-Shot Learning}
\author{
    Haozhe Zhang\textsuperscript{\rm 1}\textsuperscript{\rm 2}\equalcontrib, 
    Chenchen Jing\textsuperscript{\rm 3}\equalcontrib, 
    Mingyu Liu\textsuperscript{\rm 1}\textsuperscript{\rm 2}, 
    Qingsheng Wang\textsuperscript{\rm 4}, 
    Hao Chen\textsuperscript{\rm 1}\thanks{Corresponding Author.}
}
\affiliations{
    \textsuperscript{\rm 1}Zhejiang University,
    \textsuperscript{\rm 2}Shanghai Innovation Institute\\
    \textsuperscript{\rm 3}Zhejiang University of Technology,
    \textsuperscript{\rm 4}Northwestern Polytechnical University

    \{zjzhz,mingyuliu,haochen.cad\}@zju.edu.cn,
    jingchenchen@zjut.edu.cn,
    wqshmzh@mail.nwpu.edu.cn
    
}

\usepackage{bibentry}

\begin{document}

\maketitle

\begin{abstract}
Compositional Zero-Shot Learning (CZSL) aims to recognize unseen attribute-object compositions by learning prior knowledge of seen primitives, \textit{i.e.}, attributes and objects.
Learning generalizable compositional representations in CZSL remains challenging due to the entangled nature of attributes and objects as well as the prevalence of long-tailed distributions in real-world data.
Inspired by neuroscientific findings that imagination and perception share similar neural processes, we propose a novel approach called Debiased Feature Augmentation (DeFA) to address these challenges. The proposed DeFA integrates a disentangle-and-reconstruct framework for feature augmentation with a debiasing strategy. DeFA explicitly leverages the prior knowledge of seen attributes and objects by synthesizing high-fidelity composition features to support compositional generalization. 
Extensive experiments on three widely used datasets demonstrate that DeFA achieves state-of-the-art performance in both \textit{closed-world} and \textit{open-world} settings. 
\end{abstract}


\section{Introduction}

Humans can easily recognize a purple elephant in a science-fiction film even if they've never seen one in real life, as long as they’re familiar with purple grapes and gray elephants.
Such abilities to compose familiar primitives are known as compositional generalization \cite{fodor1988connectionism}.
Compositional Zero-Shot Learning (CZSL) \cite{misra2017red,purushwalkam2019task} is a specific setting of zero-shot learning to endow the machine with the ability of compositional generalization that recognizes unseen attribute-object compositions by learning from seen compositions. 
For example, a CZSL model trained with seen compositions such as \textit{new castle} and \textit{old bridge} should be able to correctly recognize unseen compositions such as \textit{new bridge} or \textit{old castle}. 

To generalize from seen to unseen compositions and ensure robust recognition of rare compositions, several key challenges still hinder the performance of existing CZSL models. 
Firstly, attributes and objects are often highly entangled within compositions, exhibiting complex interaction patterns. For example, given an image labeled as \textit{new castle}, it is difficult to distinguish which visual features correspond to \textit{new} and which correspond to \textit{castle}.
Secondly, the long-tailed nature of visual primitives and compositions poses significant challenges for representation learning, often leading to insufficient feature learning for rare or underrepresented patterns.


\begin{figure}[t]
\begin{center}
    \includegraphics[width=\linewidth]{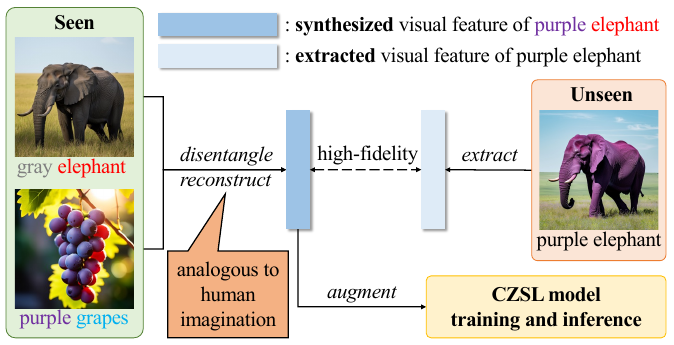}
\end{center}
\caption{
Illustration of the proposed method Debiased Feature Augmentation (DeFA). DeFA synthesizes high-fidelity composition features to augment CZSL model training and inference, analogous to how humans learn novel compositional concepts through imagination.
}
\label{fig:intro}
\end{figure}

Neuroscientific studies reveal that visual perception and imagination share overlapping neural mechanisms\cite{dijkstra2019shared}. 
Mental imagery activates an area of the cerebral cortex similar to actual visual processing, which differs mainly in signal intensity\cite{dijkstra2023subjective}.
This phenomenon suggests that humans can acquire visual concepts through both perceptual experience and internal simulation. 
Humans can establish the alignment between visual and semantic concepts of compositions even though they have never seen the compositions before, \textit{e.g.}, purple elephants. 
Humans cannot learn purple elephants by visual perception because they do not exist in reality, but can learn them by other ways such as imagination. 
It is thus desirable to synthesize composition-level features to boost compositional generalization ability for CZSL models.

Inspired by these neuroscientific findings, we propose a novel approach called \textbf{Debiased Feature Augmentation} (DeFA) in this work to address the existing challenges in CZSL research. 
DeFA integrates a disentangle-and-reconstruct framework for feature augmentation with a debiasing strategy and explicitly leverages the prior knowledge of seen attributes and objects by synthesizing high-fidelity composition features to support compositional generalization, as shown in Figure \ref{fig:intro}. 
In essence, DeFA enables the model to “imagine” rare or novel attribute-object compositions, much like how humans mentally construct unfamiliar composition concepts based on known attribute and object concepts.
The proposed approach extends the standard three-path classification pipeline \cite{yang2022decomposable, wang2023learning} by incorporating a dedicated DeFA module, which is jointly optimized with the composition classification pipeline.

Extensive experiments are conducted on three widely used CZSL datasets MIT-States \cite{mitstates}, UT-Zappos \cite{utzappos}, and C-GQA \cite{naeem2021learning}, in both \textit{closed-world} and \textit{open-world} settings. 
The experimental results demonstrate that our DeFA significantly improves the performance of the model on CZSL tasks, showing state-of-the-art performance.

The contributions of this paper are summarized as follows:
\begin{enumerate}
\item We propose a paradigm that explicitly leverages prior knowledge of seen primitives to synthesize high-fidelity compositional features to augment both model training and inference, inspired by the neuroscientific findings on the shared neural mechanisms of visual perception and imagination.
\item We propose Debiased Feature Augmentation (DeFA), a novel approach for compositional zero-shot learning.
DeFA leverages a disentangle-and-reconstruct feature augmentation framework together with a debiasing strategy to improve compositional generalization on both unseen and long-tailed compositions.
\end{enumerate}

\section{Related work}

\subsection{Compositional zero-shot learning} 
The task of compositional zero-shot learning aims to recognize unseen attribute-object compositions by learning from seen compositions. 
Existing methods mainly achieve the task via composition classification with a composition classifier \cite{misra2017red,naeem2021learning}, or primitive classifications that independently recognize attributes and objects, \cite{li2020symmetry,purushwalkam2019task}, or primitive and composition classifications for better contextuality \cite{yang2022decomposable,wang2023learning}. 
With the recent advance in pre-trained vision-language models, CLIP-based CZSL methods \cite{nayak2023learning,lu2023decomposed,huang2024troika,bao2023prompting} achieved state-of-the-art performance.
CSP \cite{nayak2023learning} first uses the CLIP \cite{radford2021learning} in CZSL. 
They replace the classes in textual prompts with trainable attributes and object tokens.
DFSP \cite{lu2023decomposed} uses a cross-modal decomposed fusion module to exploit decomposed language features in image feature learning. 
Troika \cite{huang2024troika} jointly models the vision-language alignments for the attribute, object, and composition using the CLIP. 
PLID \cite{bao2024prompting} leverages pre-trained large language models to enhance the compositionality of the softly prompted class embedding. 
The aforementioned work mainly focuses on parameter-efficient fine-tuning of CLIP.
In contrast, our method focuses on exploring the feature augmentation strategy to improve compositional generalization and only uses CLIP as a backbone.

\subsection{Feature augmentation}
Data augmentation is a prevalent training trick to improve the performance of models. 
Conventional data augmentation methods~\cite{devries2017improved,zhong2020random} usually synthesize new samples in a hand-crafted manner. 
Compared to these image-level data augmentation methods, feature augmentation is another efficient and flexible way to improve models' generalization by directly synthesizing samples in the feature space~\cite{chu2020feature,li2021simple}. 
Li et al.~\cite{li2021simple} propose a simple stochastic feature augmentation approach of perturbing feature embeddings with Gaussian noise during training, which improves the generalization ability of the model under domain change.
CFA \cite{li2023compositional} enrich the diversity of relation triplet features for debiasing in scene graph generation \cite{lu2016visual,wang2019exploring}.
In this work, we explore feature augmentation in the context of compositional zero-shot learning.
Compared to existing data augmentation applications, our method focuses on effectively leveraging prior knowledge of seen primitives to synthesize high-fidelity composition features with distribution-level coverage.

\subsection{Classification under long-tailed distribution}
Classification under long-tailed distributions is a fundamental challenge in many vision and language tasks, where the number of samples per class varies drastically. Existing methods tackle this issue by balancing the data distribution through re-sampling strategies~\cite{chawla2002smote, buda2018systematic}, or modify the loss function with cost-sensitive re-weighting~\cite{cui2019class, cao2019learning}, or modify the decision boundary by adjusting the classifier logits, either through margin-based techniques~\cite{zhang2021distribution} or class-balanced logit adjustment~\cite{menon2020long,hou2021detecting}. 
Works in \cite{chen2022zero} and \cite{jiang2024mutual} consider adjusting logits by a regulatory method involving seen and unseen classes and the presence of visual bias when re-weight samples during optimization.     
While existing methods based on resampling, loss re-weighting, or logit adjustment effectively address data imbalance, they fail to increase the intrinsic diversity of training samples. 
In this work, we enhance the data distribution by synthesizing diverse and high-fidelity samples of features through feature augmentation and integrate a frequency-aware reweighting strategy on synthesized samples to correct the bias under long-tailed distributions across attributes, objects, and compositions in the CZSL task. 

\begin{figure*}
    \begin{center}
        \includegraphics[width=0.98\textwidth]{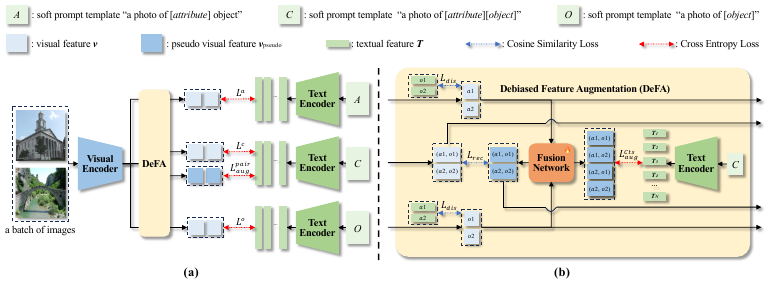}
    \end{center}
    \caption{
    (a) shows the overview of the three-path composition recognition pipeline extended with the DeFA module. 
    Given a batch of input image (the figure uses batch size 2 as an example for simplicity), the method uses a visual encoder to extract the features of compositions, attributes and objects.
    For textual inputs, text encoders extract the textual features of soft prompts constructed for
 all candidate compositions, attributes, and objects. 
    The visual and textual features are used to compute the compatibility scores for composition recognition and the cross entropy loss for model training. 
    (b) illustrates the details of the DeFA module. The fusion network synthesizes high-fidelity composition features under the constraints of disentanglement loss $\mathcal{L}_{dis}$ and reconstruction loss $\mathcal{L}_{rec}$. The DeFA module use pairwise and Cartesian feature augmentation to support model training and inference. The pairwise and Cartesian feature augmentation loss are denoted as $\mathcal{L}^{pair}_{aug}$ and $\mathcal{L}^{Cts}_{aug}$ respectively.
    }
    \label{fig:architecture}
    \end{figure*}

\section{Method}
    
    \subsection{Formulation}
    \label{subsec:formulation}
    CZSL is a task that aims at recognizing both seen and unseen attribute-object compositions during inference. 
    Given a possible attribute set $\mathcal{A} = \{ a_1, a_2, \dots, a_{|\mathcal{A}|} \}$ and a possible object set $\mathcal{O} = \{ o_1, o_2, \dots, o_{|\mathcal{O}|} \}$, the composition set $\mathcal{C} = \mathcal{A} \times \mathcal{O}$ is defined as the Cartesian product of attribute set and object set. The possible composition set $\mathcal{C}$ is also the label space of the task.
    We divide the composition set $\mathcal{C}$ into two disjoint sets, the seen set $\mathcal{C}^{s}$ and the unseen set $\mathcal{C}^{u}$, where $\mathcal{C}^{s} \cap \mathcal{C}^{u} = \emptyset$ and $\mathcal{C}^{s} \cup \mathcal{C}^{u} \subset \mathcal{C}$.     
    The training set only contains classes from the $\mathcal{C}^{s}$, while the testing set contains both seen classes and unseen classes, following the standard setting of generalized zero-shot learning \cite{pourpanah2022review,liu2021goal}. 

    Given a test image $I \in \mathcal{I}$, the CZSL model is required to predict a class label $c = (a,o)$ from the testing label set  $\mathcal{C}^{test}$.     
    In the closed-world setting, the testing label set $\mathcal{C}^{test}$ only contains the known compositions, \textit{i.e.}, $\mathcal{C}^{test} = \mathcal{C}^{s} \cup \mathcal{C}^{u}$. 
    In other words, the testing label set contains all seen classes of the training set and unseen classes of the testing set.
    In the open-world setting, the testing class set is all possible compositions, \textit{i.e.}, $\mathcal{C}^{test} = \mathcal{C}$.

    CZSL model is required to compute a compatibility score function $S: \mathcal{I} \times \mathcal{A} \times \mathcal{O} \rightarrow \mathbb{R}$ between an image $I$ and a set of candidate compositions. 
    Existing three-path classification pipelines \cite{wang2023learning,huang2023troika} usually jointly consider three kinds of compatibility score, \textit{i.e.}, the attribute compatibility score $ S^{a}(I, a)$, the object compatibility score $S^{o}(I, o)$,  and the composition compatibility score $ S^{c}(I, a, o)$. 

    \subsection{Composition recognition}
    \label{subsec:composition recognition network}
    The image encoder of the CLIP \cite{radford2021learning} is employed as the visual backbone. 
    For each input image $I$, we use the visual encoder, \textit{i.e.}, a vision transformer (ViT) \cite{dosovitskiy2021an}, to extract the feature of the [CLS] token $\bm{v} \in \mathbb{R}^{d}$, where $d$ is the dimension of feature embedding. 
    Three separate multilayer perceptrons (MLPs) are introduced to extract the attribute feature $\bm{v}^a$, the object feature $\bm{v}^o$, and the composition feature $\bm{v}^c$, respectively. 

    For textual inputs, the soft prompt \cite{nayak2023learning} is adopted to obtains the textual feature of all candidate compositions, attributes, and objects, as shown in Figure \ref{fig:architecture}. 
    The [\textit{attribute}] and [\textit{object}] tokens in the soft prompt template are learnable and initialized with the word embedding extracted by the text encoder of the CLIP. During training, each [\textit{attribute}] token and each [\textit{object}] token are mapped to the auxiliary weight vectors for the corresponding attribute and object. 
    Therefore, the textual backbone totally tunes $(|\mathcal{A}|+|\mathcal{O}|) \times d$ parameters where $d$ is the dimension of the vocabulary embedding. 
    The textual backbones obtain textual features $\bm{T}^c \in \mathbb{R}^{ N_c \times d} $, $\bm{T}^a\in \mathbb{R}^{ N_a \times d}$, $\bm{T}^o\in \mathbb{R}^{ N_o \times d}$ from three kinds of prompts. 
    The $N_c$, $N_a$, $N_o$ are the numbers of corresponding candidate compositions, attributes, and objects, respectively.
    
    We use the standard cross-entropy loss to encourage the model to recognize the composition, attribute, and object, as shown in Figure~\ref{fig:architecture} (a). 
    The compatibility scores between an image $I$ with the attribute $a$, the object $o$ and the composition $c = (a,o) $ with the aforementioned textual representations can be computed as 
    \begin{equation}
    \begin{aligned}
    \mathcal{S}^{x}(I, x) &= \operatorname{cos} (\bm{v}^x,\bm{T}^x) , \forall x \in \{a,o,c\}. 
    \end{aligned}
    \end{equation}

    The overall compatibility score for composition recognition $ \mathcal{S}_{cla}(I, c) $ is calculated as
    \begin{equation}
    \begin{aligned}
    \mathcal{S}_{cla}(I, c) = \lambda_1 \mathcal{S}^{c}(I, c)  + (1 -\lambda_1) ( \mathcal{S}^{a}(I, a) + \mathcal{S}^{o}(I, o)),
    \end{aligned}
    \end{equation}
    where $\lambda_1 $ is the hyperparameter to balance the compatibility score as well as the classification losses to be defined next. 
    
    For the baseline model, the composition $(a, o)$ with the highest score $\mathcal{S}_{cla}(I, c)$ is predicted as the label of image $I$. 
    
    The standard cross-entropy loss of a given compatibility score $\mathcal{S}^x(I,x)$ is defined as
    
    \begin{equation} \label{eq:crossentropy}
    \begin{aligned}
    {CE}(\mathcal{S}^x(I,x)) := - \log \frac{\exp\left( \mathcal{S}^x(I,x) / \tau \right)}{\sum\limits_{x' \in \mathcal{X}} \exp\left( \mathcal{S}^{x}(I, x') / \tau \right)}, \\
    \forall x \in \{a, o, c\},\ \mathcal{X} = \begin{cases}
    \mathcal{A}, & \text{if } x = a \\
    \mathcal{O}, & \text{if } x = o \\
    \mathcal{C}, & \text{if } x = c 
    \end{cases},
    \end{aligned}
    \end{equation}
    where $\tau \in \mathbb{R}$ is the pre-defined temperature parameter of CLIP.
    
    The three-path classification losses are further calculated as 
    \begin{equation} \label{eq:loss_classification}
    \begin{aligned}
    \mathcal{L}^x = {CE}(\mathcal{S}^x(I,x)), \forall x \in \{a,o,c\}.
    \end{aligned}
    \end{equation}

    The overall classification loss is calculated as
    \begin{equation}
    \begin{aligned} \label{eq:overall_loss}
    \mathcal{L}_{cla} = \lambda_1 \mathcal{L}^c + (1 - \lambda_1) (\mathcal{L}^a + \mathcal{L}^o).
    \end{aligned}
    \end{equation}

    \subsection{Disentanglement and reconstruction}
    \label{subsec:Composition feature synthesis}

    We adopt the disentanglement loss \cite{jing2024retrieval} to promote the disentanglement between attributes and objects. The disentanglement loss penalizes the cosine similarity between the attribute feature and the ground truth object label feature, as well as the cosine similarity between the object feature and the ground truth attribute label feature. The disentanglement loss is computed as follows
    \begin{equation}
    \begin{aligned}
    \mathcal{L}_{dis}= \operatorname{cos} ( \bm{v}^a , \bm{T}^o_{gt})  + \operatorname{cos} (\bm{v}^o, \bm{T}^a_{gt}) ,
    \end{aligned}
    \end{equation}
    where $ \bm{T}^o_{gt}$ and $ \bm{T}^a_{gt} $ are the textual features of the ground truth object label and the ground truth attribute label, respectively. 
    
    
    
    
    Considering attributes and objects are often entangled within compositions with complex interaction patterns, we proposed a fusion network to learn the synthesizing pattern of pseudo samples of composition features. The fusion network is a trainable network with residual connection, as shown in Figure~\ref{fig:architecture} (b). The input of the fusion network is formed by concatenating the attribute feature $\bm{v}^a$ and object feature $\bm{v}^o$, while the output has the same dimensionality as the real composition feature $\bm{v}^c$. 
    The fusion network can be formulated as follows
    \begin{equation}
    \label{eq:fusion_network}
    \begin{aligned}
    \bm{v}^{c}_{pseudo}=\alpha\mathcal{F}_{\theta}(\bm{v}^a,\bm{v}^o)+(1-\alpha)(\bm{v}^a+\bm{v}^o),
    \end{aligned}
    \end{equation}
    where $\alpha$ is a hyperparameter to balance the residual connection. $\mathcal{F}_{\theta}$ denotes a trainable network and in this work is an MLP. $ \bm{v}^{c}_{pseudo}$is the pseudo sample of the composition feature. 
    
    The fusion network aims to acquire a generalizable mechanism for integrating attribute and object features, enabling effective transfer to synthesize various compositions, including unseen and long-tailed compositions. 
    
    We introduce reconstruction loss as a supervisory signal to ensure that the composition feature synthesized by the fusion network approximates the real composition feature as closely as possible. 
    
    The reconstruction loss is the negative value of cosine similarity between the synthesized pseudo sample and real sample. The reconstruction loss is calculated as follows.
    \begin{equation} 
    \label{eq:resconstruction_loss}
    \begin{aligned} 
    \mathcal{L}_{rec}= - \operatorname{cos} ( \bm{v}^{c}_{pseudo} , \bm{v}^{c}),
    \end{aligned}
    \end{equation}

    As shown on the left side of the fusion network in Figure~\ref{fig:architecture} (b), we concatenate the attribute features and the object features in the original corresponding order in the batch and feed them into the fusion network. The output on the left side of the fusion network in Figure~\ref{fig:architecture} (b) and the composition feature extracted by the MLP of the composition subbranch are pair-to-pair corresponding to the label. Thus, we can calculate the reconstruction loss between the synthesized pseudo samples and corresponding real samples according to Eq.~\ref{eq:resconstruction_loss}. The reconstruction loss guides the fusion network to learn the correct composition feature fusion paradigm, thereby enabling the fusion network to synthesize effective pseudo samples. 



    \subsection{Debiased feature augmentation}
    \label{subsec:Debiased feature augmentation}
    

    Given a batch $\mathcal{B}$ of size $|\mathcal{B}|$, we denote the attribute labels as $\mathcal{B}^a = \{ a_{1}, a_{2}, \dots, a_{|\mathcal{B}|} \}$, the object labels as $\mathcal{B}^o = \{ o_{1}, o_{2}, \dots, o_{|\mathcal{B}|} \}$, and the composition labels as $\mathcal{B}^c = \{ (a_{1},o_{1}), (a_{2},o_{2}), \dots, (a_{|\mathcal{B}|},o_{|\mathcal{B}|}) \}$.

    In the batch $\mathcal{B}$, we can obtain a batch of attribute visual features $\mathcal{V}^a = \{ \bm{v}^a_{1}, \bm{v}^a_{2}, \dots, \bm{v}^a_{|\mathcal{B}|} \}$ and a batch of object visual features $\mathcal{V}^o = \{ \bm{v}^o_{1}, \bm{v}^o_{2}, \dots, \bm{v}^o_{|\mathcal{B}|} \}$. We perform the Cartesian product on $\mathcal{V}^a$ and $\mathcal{V}^o$ and concatenate the elements of the two sets pairwise. As a result, we can get a set of concatenated visual features, defined as $\mathcal{V}_{concat} = \mathcal{V}^a \times \mathcal{V}^o$. 
    The $\mathcal{V}_{concat}$ is fed into the fusion network and we can obtain pseudo composition features for feature augmentation, defined as $\mathcal{V}_{pseudo}$. 
    
    For textual inputs in DeFA module, we adopt the prompt template "a photo of [\textit{attribute}][\textit{object}]", in which the [\textit{attribute}] token and the [\textit{object}] token are replaced with the corresponding token of attribute or object in $\mathcal{V}_{pseudo}$. The text encoder in the DeFA module extracts the textual features from the concatenated prompts and builds a set of corresponding pseudo textual features, denotes as $\bm{T}^c_{pseudo}$.
    In Figure~\ref{fig:architecture} (b), $T_i,i \in\{1,2,\dots,N\}$ denotes $\bm{T}^c_{pseudo}$ and $N=|\mathcal{A}| \times |\mathcal{O}|$. 
    
    Note that the [\textit{attribute}] token and the [\textit{object}] token in the DeFA module are map to the same auxiliary weight vectors as mentioned in \ref{subsec:composition recognition network}, that is, the textual backbones in the whole model share the learnable $(|\mathcal{A}|+|\mathcal{O}|) \times d$ parameters. 

    The pairwise feature augmentation score, denoted as $\mathcal{S}^{pair}_{aug}$, measures the compatibility between a pseudo visual feature and its pairwise corresponding ground truth composition within the batch. It is computed as
    \begin{equation}
    \begin{aligned}
    \mathcal{S}^{pair}_{aug}(\bm{v}^c_{pseudo}, a, o) &=  \operatorname{cos} (\bm{v}^c_{pseudo}, \bm{T}^c_{gt}), \forall (a,o) \in \mathcal{B}^c,
    \end{aligned}
    \end{equation}
    where $\bm{T}^c_{\text{gt}}$ denotes the ground truth textual feature pairwise corresponding to the composition label $(a, o)$ from $\mathcal{B}^c$. 

    The pairwise feature augmentation loss, denoted as $\mathcal{L}^{pair}_{aug}$, is formulated as a cross-entropy loss based on $\mathcal{S}^{pair}_{aug}$, encouraging the pseudo visual feature to align with its pairwise corresponding ground truth textual composition in the batch. $\mathcal{L}^{pair}_{aug}$ is calculated as
    \begin{equation} \label{eq:loss_fa}
    \begin{aligned}
    \mathcal{L}^{pair}_{aug} = CE(\mathcal{S}^{pair}_{aug}(\bm{v}^c_{pseudo}, a, o)).
    \end{aligned}
    \end{equation}
    
    During inference, the compatibility score for composition recognition and pairwise feature augmentation scores are combined to complement the composition recognition.
    
    
    The overall compatibility score $ \mathcal{S}(I, a, o) $ augmented by $\mathcal{S}^{pair}_{aug}$ is calculated as
    \begin{equation}
    \begin{aligned}
    \mathcal{S}(I, a, o) = \beta \mathcal{S}_{cla}(I, a, o) + (1-\beta) \mathcal{S}^{pair}_{aug}(\bm{v}^c_{pseudo}, a, o), 
    \end{aligned}
    \end{equation}
    
    Here $\beta$ is a hyperparameter to balance the scores. For our model, the composition $(a, o)$ with the highest score $\mathcal{S}(I, a, o)$ is predicted as the label of image $I$. 

    In each training batch, attribute features and object features are concatenated via the Cartesian product and then fed into the fusion network, as shown in the right side of the fusion network in Figure~\ref{fig:architecture} (b). 

    The Cartesian feature augmentation score, denoted as $\mathcal{S}^{Cts}_{aug}$, measures the compatibility between a pseudo visual feature and all candidate attribute-object compositions constructed from the Cartesian product of the current batch of attribute features and object features, \textit{i.e.}, from $\mathcal{B}^a \times \mathcal{B}^o$. $\mathcal{S}^{Cts}_{aug}$ is computed as

    \begin{equation}
    \begin{aligned}
    \mathcal{S}^{Cts}_{aug}(\bm{v}^c_{pseudo}, a, o) =  \operatorname{cos} (\bm{v}^c_{pseudo}, \bm{T}^c_{pseudo}), \\ \forall (a,o) \in \mathcal{B}^a \times \mathcal{B}^o.
    \end{aligned}
    \end{equation}

    To alleviate the impact of long-tailed distribution, \textit{i.e.}, the data imbalance issue across attributes, objects, and compositions, we adopt a frequency-aware debiasing weight scheme for feature augmentation. Specifically, we compute the weights $w^a_{de}$, $w^o_{de}$, $w^c_{de}$ for attributes, objects, and compositions based on their occurrence statistics in the training set. The weight for each concept is defined as

    \begin{equation}
    \begin{aligned}
    w^x_{de} = \frac{1 / (|\mathcal{K}^x| + 1)^{\rho}}{\sum_{i=1}^{|\mathcal{X}|} \left(1 / (|\mathcal{K}^x_i| + 1)^{\rho} \right)} \cdot |\mathcal{X}|,\\ 
    \forall x \in \{a, o, c\},\ \mathcal{X} = \begin{cases}
    \mathcal{A}, & \text{if } x = a \\
    \mathcal{O}, & \text{if } x = o \\
    \mathcal{C}, & \text{if } x = c 
    \end{cases}.
    \end{aligned}
    \end{equation}

    Here, $|\mathcal{K}^a|$, $|\mathcal{K}^o|$ and $|\mathcal{K}^c|$ denote the frequency of a given attribute, object, and composition in the training data, respectively. 
    The constant $+1$ avoids division by zero and stabilizes the scaling for rare categories. The hyperparameter $\rho \in [0,1]$ controls the strength of frequency suppression:
    When $\rho = 0$, all categories are treated equally;
    When $\rho = 1$, the weights follow strict inverse-frequency scaling;
    Intermediate values (e.g., $\rho = 0.5$) provide softened debiasing. 
    
    
    To comprehensively account for the long-tailed distributions across attributes, objects, and compositions, we construct a overall debiasing weight by combining their individual frequency-aware weights. The overall debiasing weight is formulated as
    \begin{equation}
    \begin{aligned}
    w_{de} = \mu w^c_{de} + (1 - \mu)(w^a_{de} + w^o_{de}), 
    \end{aligned}
    \end{equation}
    where $\mu \in [0,1]$ is a balancing hyperparameter that controls the trade-off between composition-level weighting and the primitive-level (attribute and object level) weighting. This enables the model to mitigate distributional bias at both the primitive level and the composition level.
    

    The Cartesian feature augmentation loss, denoted as $\mathcal{L}^{Cts}_{aug}$, is a weighted cross entropy loss based on the debiasing weight and the Cartesian feature augmentation score. The Cartesian feature augmentation loss is calculated as
    \begin{equation} \label{eq:loss_fa}
    \begin{aligned}
    \mathcal{L}^{Cts}_{aug} = w_{de} \cdot CE(\mathcal{S}^{Cts}_{aug}(\bm{v}^c_{pseudo}, a, o)).
    \end{aligned}
    \end{equation}
    
    We jointly optimize the composition recognition network and the DeFA module. The overall loss for the optimization is given by 
    \begin{equation}
    \begin{aligned} \label{eq:overall_loss}
    \mathcal{L} = \mathcal{L}_{cla} + \lambda_{2}\mathcal{L}_{dis} + \lambda_{3}\mathcal{L}_{rec} + \lambda_{4}\mathcal{L}^{pair}_{aug} + \lambda_{5}{\mathcal{L}^{Cts}_{aug}},
    \end{aligned}
    \end{equation}
    where $\lambda_2$, $\lambda_3$, $\lambda_4$, and $\lambda_5$ are hyperparameters to balance the losses. 

\begin{table}[t]
\centering
{
\setlength{\tabcolsep}{1mm}
\begin{tabular}{lrrr rr rrr}
\toprule
 &  & & & \multicolumn{1}{c}{\textbf{Train}} & \multicolumn{2}{c}{\textbf{Val}} & \multicolumn{2}{c}{\textbf{Test}} \\
\textbf{Dataset}& {$|\mathcal{I}|$} & {$|\mathcal{A}|$} & {$|\mathcal{O}|$} & {$|\mathcal{C}^{s}|$} & {$|\mathcal{C}^{s}|$} & {$|\mathcal{C}^{u}|$} & {$|\mathcal{C}^{s}|$} & {$|\mathcal{C}^{u}|$}  \\
\hline
MIT-States &53k & 115 & 245 & 1262  & 300 & 300  & 400 & 400  \\
UT-Zappos &30k& 16  & 12  & 83 & 15  & 15 & 18  & 18  \\
C-GQA &39k& 413 & 674 & 5592 & 1252 & 1040 & 888 & 923  \\
\bottomrule
\end{tabular}
}
\caption{Statistics of the MIT-States, the UT-Zappos, and the C-GQA.}
\label{tab:exp-dataset}
\end{table}

\begin{table*}[t]
    \caption{
    Quantitative results of the proposed method and the state-of-the-art on CZSL datasets in the \textit{closed-world} setting. 
    } 
    \centering
    
    {%
    \setlength{\tabcolsep}{1mm}
    \small
    \begin{tabular}{ll|cccc|cccc|cccc}
    \toprule
    \multicolumn{2}{c|}{\textit{\textbf{Closed-world}}} & \multicolumn{4}{c|}{ \textbf{MIT-States}}          & \multicolumn{4}{c|}{\textbf{UT-Zappos}}                & \multicolumn{4}{c}{\textbf{C-GQA}}            \\
    Method & Venue & AUC  & HM   & Seen     & Unseen  & AUC  & HM   & Seen     & Unseen  & AUC  & HM   & Seen     & Unseen  \\ 
    \hline
    CSP \cite{nayak2023learning} & ICLR'23  & 19.4 & 36.3 & 46.6 & 49.9 & 33.0& 46.6& 64.2& 66.2& 6.2& 20.5& 28.8& 26.8\\
    DFSP \cite{lu2023decomposed} & CVPR'23 & 20.6 & 37.3 & 46.9 & 52.0 & 36.9& 47.2& 66.7& 71.7& 10.5& 27.1& 38.2& 32.9\\
    DLM \cite{hu2024dynamic} & AAAI'24 & 20.0 & 37.4  & 46.3 &	49.8   & 39.6& 52.0& 67.1&	72.5& 7.3& 21.9& 32.4&	28.5\\
    ProLT \cite{jiang2024revealing} & AAAI'24 & 21.1 & 38.2  & 49.1 & 51.0 &  36.1& 49.4& 66.0& 70.1& 11.0& 27.7& 39.5&	32.9\\
    Troika \cite{huang2023troika} & CVPR'24 & 22.1 & \textbf{39.3} & 49.0 & \textbf{53.0}  & 41.7& 54.6& 66.8& 73.8& 12.4& 29.4& 41.0& 35.7\\
    PLID \cite{bao2024prompting}  & ECCV'24 & 22.1 & 39.0 & 49.7 & 52.4  & 38.7& 52.4& 67.3& 68.8& 11.0& 27.9& 41.0& \textbf{38.8}\\
    \hline
    \multicolumn{2}{c|}{DeFA(ours)} & \textbf{22.8} & \textbf{39.3}& \textbf{51.1}& 52.9& \textbf{46.1}& \textbf{58.6}& \textbf{67.9}& \textbf{75.9}& \textbf{14.6}& \textbf{32.3}& \textbf{45.2}& 37.2\\ 
    \bottomrule
    \end{tabular}%
    }
    \label{tab:closed}
    \end{table*}

    \begin{table*}[t]
    \caption{
    Quantitative results of the proposed method and the state-of-the-art on CZSL datasets in the \textit{open-world} setting. 
    } 
    \centering
    {

    \setlength{\tabcolsep}{1mm}
    \small
    \begin{tabular}{ll|cccc|cccc|cccc}
    \toprule
     \multicolumn{2}{c|}{\textit{\textbf{Open-world}}} & \multicolumn{4}{c|}{ \textbf{MIT-States}}          & \multicolumn{4}{c|}{\textbf{UT-Zappos}}                & \multicolumn{4}{c}{\textbf{C-GQA}}            \\
    Method & Venue & AUC  & HM   & Seen     & Unseen  & AUC  & HM   & Seen     & Unseen & AUC  & HM   & Seen     & Unseen  \\ 
    \hline
     CSP \cite{nayak2023learning}  & ICLR'23 & 5.7 & 17.4 & 46.3 & 15.7 & 22.7& 38.9& 64.1& 44.1& 1.2& 6.9& 28.7& 5.2\\ 
    DFSP \cite{lu2023decomposed} & CVPR'23 & 6.8 & 19.3 & 47.5 & 18.5 & 30.3& 44.0& 66.8& 60.0& 2.4& 10.4& 38.3& 7.2\\
    Troika \cite{huang2023troika} & CVPR'24 & 7.2 & 20.1 & 48.8 & 18.7 & 33.0& 47.8& 66.4& 61.2& 2.7& 10.9& 40.8& 7.9\\    
    PILD \cite{bao2024prompting} & ECCV'24 & 7.3 & 20.4 & 49.1 & 18.7 & 30.8& 46.6& 67.6& 55.5& 2.5& 10.6& 39.1& 7.5\\   
    \hline
    \multicolumn{2}{c|}{DeFA(ours)} & \textbf{8.3} & \textbf{21.7} & \textbf{51.1} & \textbf{20.1} & \textbf{35.5} & \textbf{50.9} & \textbf{68.0} & \textbf{63.1} & \textbf{4.1} & \textbf{14.2} & \textbf{45.2} & \textbf{10.7} \\
    \bottomrule
    \end{tabular}%
    }
    \label{tab:open}
    \end{table*}

\section{Experiment}


\subsection{Experimental setup}
\noindent \textbf{Dataset.} 
We benchmark our method on three established Compositional Zero-Shot Learning datasets: MIT-States \cite{mitstates}, UT-Zappos \cite{utzappos}, and C-GQA \cite{naeem2021learning}. 
The split suggested by previous works \cite{purushwalkam2019task,nayak2023learning} is adopted. 
The detailed dataset statistics are summarized in Table \ref{tab:exp-dataset}.

\noindent \textbf{Metrics. } 
Following the compositional zero-shot learning (CZSL) evaluation protocol, we adopt the calibration strategy proposed by~\cite{purushwalkam2019task}, including best seen composition accuracy (\textbf{Seen}), best unseen class accuracy (\textbf{Unseen}), best harmonic mean (\textbf{HM}) between all seen and unseen accuracies, and area under a curve (\textbf{AUC}) with all seen and unseen accuracies as horizontal and vertical axes.
The best-case accuracy demonstrates the potential of the model in extreme calibration settings, the best harmonic mean balances accuracy trade-offs, and the area under a curve quantifies overall performance of a model.
We evaluate our model in both \textit{closed-world} and \textit{open-world} setting. In the \textit{open-world} setting, the GloVe model \cite{pennington2014glove} is employed to perform feasibility calibration, thereby filtering out infeasible compositions.

    \begin{figure}[t]
    \begin{center}
        \includegraphics[width=\linewidth]{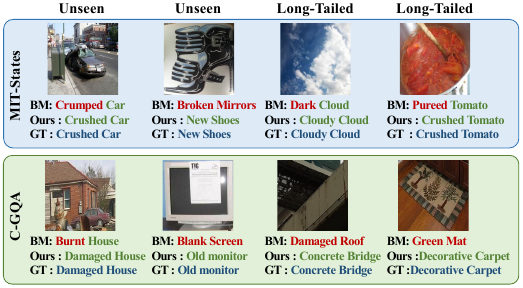}
    \end{center}
    \caption{
    Qualitative results of the proposed method on the MIT-States and the C-GQA.
    For each sample, we show an input image, predictions of the baseline model (BM) and our model (Ours), and the ground truth (GT).
    }
    \label{fig:case}
    \end{figure}


    \subsection{Implementation details}
    \noindent \textbf{Model. }
    We implement our method using PyTorch and conduct all training and evaluation on NVIDIA RTX 4090 GPU.
    Following prior works \cite{lu2023decomposed, huang2023troika}, We adopt the image and text encoders from CLIP with the ViT-L/14 backbone as our visual and text encoders. We use three two-layer MLPs for the MIT-States and C-GQA datasets, and three one-layer self-attention blocks for UT-Zappos as visual feature projectors \cite{jing2024retrieval} in the composition recognition network.
    The representation dimensionality is fixed at $d = 1024$ across all three datasets. 
    We apply LoRA \cite{hu2021lora}, a lightweight and parameter-efficient fine-tuning strategy, to the top $2$/$2$/$12$ layers of the CLIP image encoder for MIT-States, UT-Zappos, and C-GQA, respectively.
    To encode textual inputs, we adopt soft prompts \cite{nayak2023learning} to obtain representations for all candidate compositions, attributes, and objects. 
    The trainable fusion network in the DeFA module adopts a three-layer MLP for both MIT-States and C-GQA, and adopts an one-layer MLP for UT-Zappos. 

    \noindent \textbf{Training setup. } 
    The number of training epochs is set to $20$ for both UT-Zappos and C-GQA, and to $50$ for MIT-States.
    The batch size during training is set to $128$ for both UT-Zappos and MIT-States, and to $32$ for C-GQA.
    The hyperparameter $\alpha$, which balances the residual connection in the fusion network, is set to $0.8$/$0.8$/$0.9$ for MIT-States, UT-Zappos, and C-GQA, respectively.
    The hyperparameter $\beta$, used to balance the basic composition recognition score and the pairwise feature augmentation score, is fixed at $0.5$ for all three datasets.
    The hyperparameter $\rho$, which controls the strength of frequency debiasing, is fixed at $0.5$ for all three datasets.
    The hyperparameter $\mu$, which controls the trade-off between composition-level debiasing and the primitive-level debiasing, is set to $0.9$/$0.8$/$0.9$ for MIT-States, UT-Zappos, and C-GQA, respectively. 
    For the MIT-States, the hyperparameters $\lambda_1 $, $\lambda_2$, $\lambda_3$, $\lambda_4$ and $\lambda_5$ to balance the losses are set as $0.3$, $10.0$, $100.0$, $0.7$ and $0.1$.
    For UT-Zappos, they are set as $0.9$, $10.0$, $10.0$, $0.9$ and $0.1$.
    For C-GQA, they are set as $0.2$, $3.0$, $1.0$, $0.1$ and $0.1$.
    We present the results of the hyperparameter sensitivity analysis in the supplementary material.

    \subsection{Main results}
    \noindent \textbf{Quantitative results. }
    We compare our method with various state-of-the-art methods. 
    The quantitative results of the proposed method and the state-of-the-art methods on the test split of CZSL datasets in the \textit{closed-world} setting and in the \textit{open-world} setting are shown in Table \ref{tab:closed} and Table \ref{tab:open}, respectively. 
    The best results of methods are shown in bold. 
    In the closed-world setting, we observe that our method outperforms all other methods on three datasets. 
    The main reason is that, benefiting from DeFA, the model can effectively leverage the prior knowledge of seen primitives to synthesize high-fidelity and diverse features, including long-tailed pairs, to augment training and inference. We also compare the classification accuracy between the baseline model and DeFA. 
    The classification accuracy on the bottom 80\% least frequent categories in the training set of the baseline model on UT-Zappos is 21.1\%, while the classification accuracy of the full model is 32.8\%. This demonstrates improved generalization on long-tailed pairs.

    In the challenging \textit{open-world} setting, our method is state-of-the-art on the three datasets, especially on the area under a curve and best unseen class accuracy metrics. 
    In the \textit{open-world} setting, all possible categories should be considered. Benefiting from DeFA, our method allows the model learn the knowledge of unseen pairs by synthesizing high-fidelity features of them, analogous to how humans learn novel compositional concepts through imagination.
    Note that we use identical model weights for both closed-world and open-world setting. 
    
    \noindent \textbf{Qualitative results. }
    We also provide qualitative examples from the MIT-States and the C-GQA in Figure~\ref{fig:case}.
    We focus on challenging cases, \textit{i.e.}, unseen or long-tailed compositions. 
    For each test image, we present the input image along with the predictions from the baseline model and our full model, as well as the ground truth. 
    Our model demonstrates the ability to accurately recognize some challenging long-tailed attributes, objects, and compositions. In contrast, the baseline model tends to predict categories that occur more frequently from the training set for long-tailed labels. Benefiting from debiased feature augmentation, our model exhibits increased confidence in predicting long-tailed attributes, objects, and compositions.

    \begin{table}[t]
    \caption{Results of different variants of our model on the UT-Zappos dataset.}
    \label{tab:exp-ablation}
    \centering
    \begin{tabular}{c|c|cccc}
        \toprule
        No. & Model Type & AUC & HM & Seen & Unseen \\
        \hline
        1 & baseline model & 41.2 & 52.0 & 67.2 & 72.3 \\
        2 & w/o $\mathcal{L}_{rec}$ & 40.5 & 54.8 & 65.7 & 71.9 \\
        3 & w/o $\mathcal{L}^{pair}_{aug}$ & 40.6 & 55.1 & 64.4 & 71.3 \\
        4 & w/o $\mathcal{L}^{Cts}_{aug}$ & 43.1 & 55.7 & 66.7 & 74.7 \\
        5 & w/o $\mathcal{F}_{\theta}$ & 40.0 & 53.4 & 67.7 & 71.7 \\
        \hline
        6 & ours & \textbf{46.1} & \textbf{58.6} & \textbf{67.9} & \textbf{75.9} \\
        \bottomrule
    \end{tabular}
    \end{table}
    \subsection{Ablation study}
    In this section, we ablate the proposed DeFA module to evaluate the effectiveness of its key components. Table~\ref{tab:exp-ablation} reports the performance of different variants of our model on the UT-Zappos dataset in the \textit{closed-world} setting. Note that the baseline model uses the same image encoder of CLIP tuned with LoRA as the full model. 
    
    Ablating the proposed reconstruction loss $\mathcal{L}_{rec}$ leads to a notable drop in AUC 5.6 points (12.1\%) on UT-Zappos, highlighting its role as an effective supervision signal for feature reconstruction.
    Removing $\mathcal{L}^{pair}_{aug}$ or $\mathcal{L}^{Cts}_{aug}$ results in significant performance degradation, indicating that both losses contribute to learning from the synthesized features.
    The variant of the model without $\mathcal{F}_{\theta}$ specifically refers to the model without the learnable network in the DeFA module. According to Eq. \ref{eq:fusion_network}, in this variant, we directly add the attribute feature and the object feature as the pseudo feature. The AUC of this variant dropped by $6.1$ points ($13.2\%$), and the best unseen class accuracy dropped by $4.2$ points ($5.5\%$), highlighting that the learnable fusion network plays a crucial role in boosting compositional generalization by synthesizing high-fidelity features for unseen compositions.

    \begin{figure}[t]
    \centering
    \includegraphics[width=\linewidth]{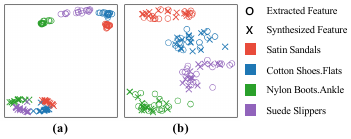}
    \caption{Visualization of the unseen composition feature distribution of model w/o $\mathcal{L}_{rec}$ (a) and full model (b). }
    \label{fig:t-SNE}
    \end{figure}
    \subsection{Visualization of Feature Distribution}


    We visualize the distribution with the t-SNE tool \cite{van2008visualizing} in Figure~\ref{fig:t-SNE} to demonstrate the effectiveness of the DeFA in synthesizing high-fidelity composition features with distribution-level coverage.
    We compared our full model with the model without $\mathcal{L}_{rec}$ to show the effectiveness of $\mathcal{L}_{rec}$ in supervising feature synthesis. We select four unseen compositions in the test split of UT-Zappos and choose $16$ images for each composition. Figure~\ref{fig:t-SNE} (a) shows the results of the model without $\mathcal{L}_{rec}$ and (b) shows the results of our full model. 
    The circle denotes the extracted feature, while the cross denotes the synthesized feature. 
    Each circle or cross refers to an image, and different colors denote different compositions. 
    
    In contrast, we observe that our full model is capable of synthesizing unseen composition features that are closely aligned with the real extracted features at the distribution level, while the model without $\mathcal{L}_{rec}$ fails to do so. 
    This confirms that $\mathcal{L}_{rec}$ plays a critical role in guiding the fusion network in learning a generalizable fusion paradigm that effectively integrates attribute and object features, enabling it to synthesize high-fidelity features for unseen compositions.



\section{Conclusion}
Inspired by the similar neural mechanisms underlying human visual perception and imagination, we propose DeFA to tackle key challenges in CZSL. DeFA explicitly leverages the prior knowledge of attributes and objects to synthesize high-fidelity compositional features via a disentangle-and-reconstruct paradigm. This enables the model to simulate a wide range of compositions, including both unseen and long-tailed ones. This paradigm mirrors the human ability to learn unseen or infrequent visual concepts by imagining them.
To mitigate the effects of long-tailed data distributions, we further introduce a debiased training strategy that encourages the model to focus on underrepresented compositions. 
Extensive experiments demonstrate that DeFA achieves state-of-the-art performance, thereby validating the effectiveness and robustness of our approach.

\bibliography{aaai2026}

\end{document}